\documentclass[conference]{IEEEtran}

\ifCLASSINFOpdf
  \usepackage[pdftex]{graphicx}
  \DeclareGraphicsExtensions{.pdf,.jpeg,.png}
\else
\fi

\usepackage{amsfonts, graphicx,amsmath,amsbsy,amssymb}

\newcommand{\Mc}{{\mathcal M}}
\newcommand{\Nc}{{\mathcal N}}

\newcommand{\Pc}{{\mathcal P}}
\newcommand{\Qc}{{\mathcal Q}}
\newcommand{\Rc}{{\mathcal R}}

\newcommand{\Wc}{{\mathcal W}}

\newcommand{\Rd}{{\mathbb R}}

\begin{document}
%
\title{Deep Learning Reconstruction  for 9-View  Dual Energy CT Baggage Scanner}

\author{
\IEEEauthorblockN{Yoseob Han}
\IEEEauthorblockA{
KAIST, Daejeon, Korea\\
Email: hanyoseob@kaist.ac.kr}
\and
\IEEEauthorblockN{Jingu Kang}
\IEEEauthorblockA{
GEMSS Medical Co. Seongnam, Korea\\
Email:  jingu.kang@gemss-medical.com}
\and
\IEEEauthorblockN{Jong Chul Ye}
\IEEEauthorblockA{
KAIST, Daejeon, Korea\\
Email: jong.ye@kaist.ac.kr}}



\maketitle

\begin{abstract}
For homeland and transportation  security applications, 
2D X-ray explosive detection system (EDS)  have been widely used, but  they have limitations in recognizing 3D shape of the hidden objects.
Among various types of  3D computed tomography (CT)  systems to address this issue,
this paper is interested in a
stationary CT  using
fixed X-ray sources and detectors.
 However, due to the limited number of projection views, analytic reconstruction 
 algorithms produce severe streaking artifacts.
  Inspired by recent success of deep learning approach for sparse view CT reconstruction,
 here we propose a novel image and sinogram domain deep learning architecture
 for 3D reconstruction from very sparse view measurement.
 The algorithm has been tested with the real data from a prototype 9-view dual energy stationary CT EDS carry-on baggage scanner  developed
 by GEMSS Medical Systems,  Korea, which confirms the superior reconstruction performance
 over the existing approaches.
 \end{abstract}


\IEEEpeerreviewmaketitle

\section{Introduction}
\label{sec:intro}

In homeland and aviation security applications, there has been increasing demand 
for  X-ray CT EDS system for carry-on baggage screening. 
A CT-EDS can produce an accurate 3D object structure for segmentation and threat detection, which is often not possible when a 2D-EDS system captures projection views in only one or two angular directions.
There are currently two types of CT EDS systems: gantry-based CT and stationary CT.
 While  gantry-based CT EDS is largely the same as  medical CT,  baggage screening
 should be carried out continuously, so it is often difficult to  continuously screen carry-on bags because of  the
 possible mechanical overloading of the gantry system.
 On the other hand, a stationary CT EDS system uses  fixed X-ray sources and detectors,
making the system suitable for routine
 carry-on baggage inspection.
 
 \begin{figure}[!hbt]
  \centering
  \centerline{\includegraphics[width = 0.45\textwidth]{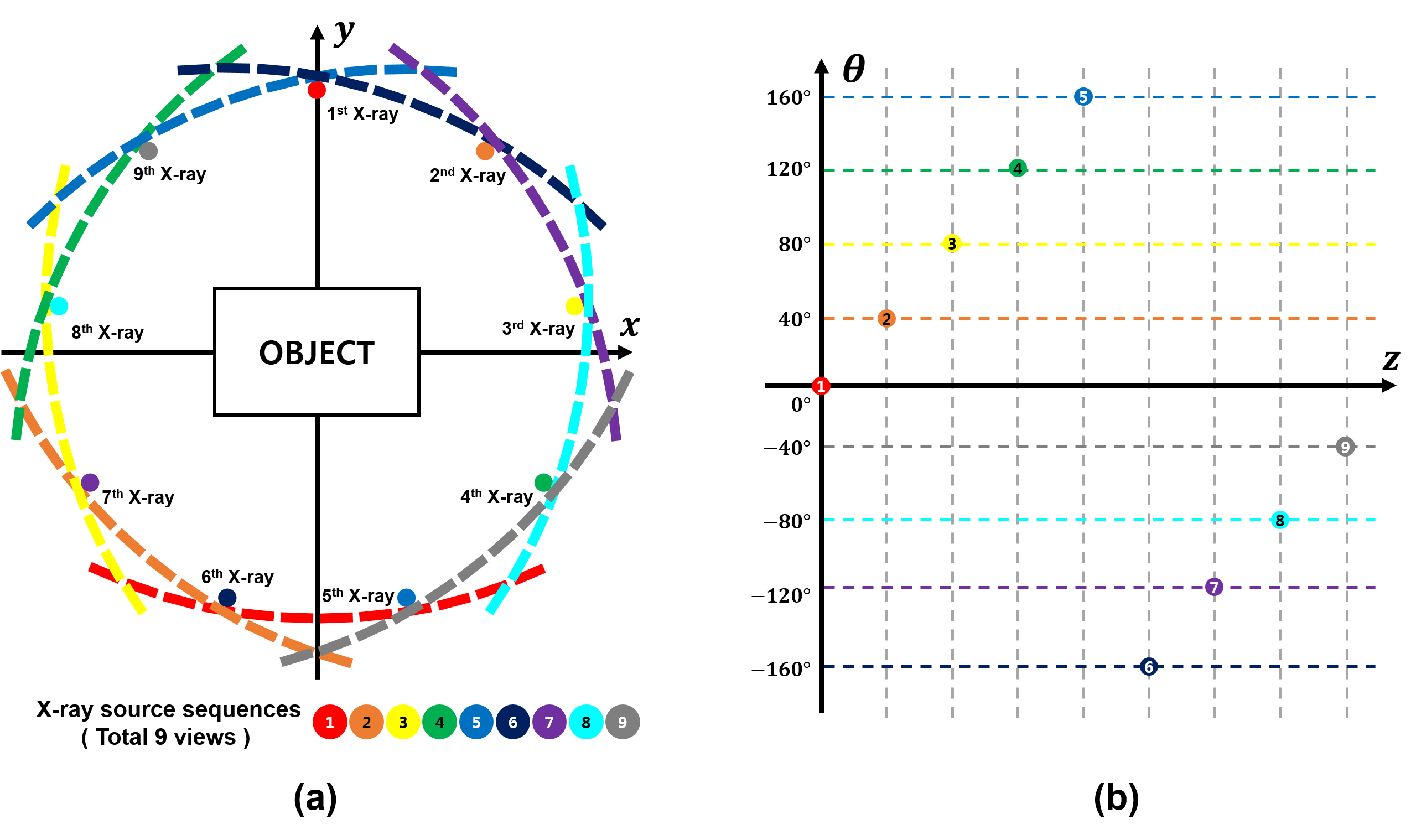}}
  \centerline{}\medskip
\vspace{-1cm}
\caption{X-ray source positions in our prototype 9 view dual energy CT EDS: (a) $x-y$  direction and  (b)  $\theta-z$ direction, respectively. 
 }\label{fig:ct_system} 
\end{figure}
 
For example,  Fig. \ref{fig:ct_system} shows source and detector geometry of  the prototype stationary CT-EDS system developed by GEMSS Medical Systems, Korea. 
As shown in Fig. \ref{fig:ct_system}(a),  nine  pairs of X-ray source and dual energy detector in  the opposite direction are distributed at the same angular interval.
For seamless screening without stopping convey belt,  each pair of source and detectors are arranged along the z-direction   as shown in Fig. \ref{fig:ct_system}(b) so that different projection view 
data can be collected while   the carry-on baggages moves  continuously on the conveyor belt. 
Then,  9-view fan beam projection data  is obtained for each z-slice by rebinning the measurement data.
This type of stationary CT system is suitable for EDS applications because it does not require a rotating gantry, but  with only 9 projection views
it is difficult to use a conventional  filtered backprojection (FBP) algorithm  due to 
 severe streaking artifacts.
 Therefore, advanced reconstruction algorithms with fast reconstruction time are required.

\begin{figure*}[!tb]
  \centering
  \centerline{\includegraphics[width =15cm,height=5cm]{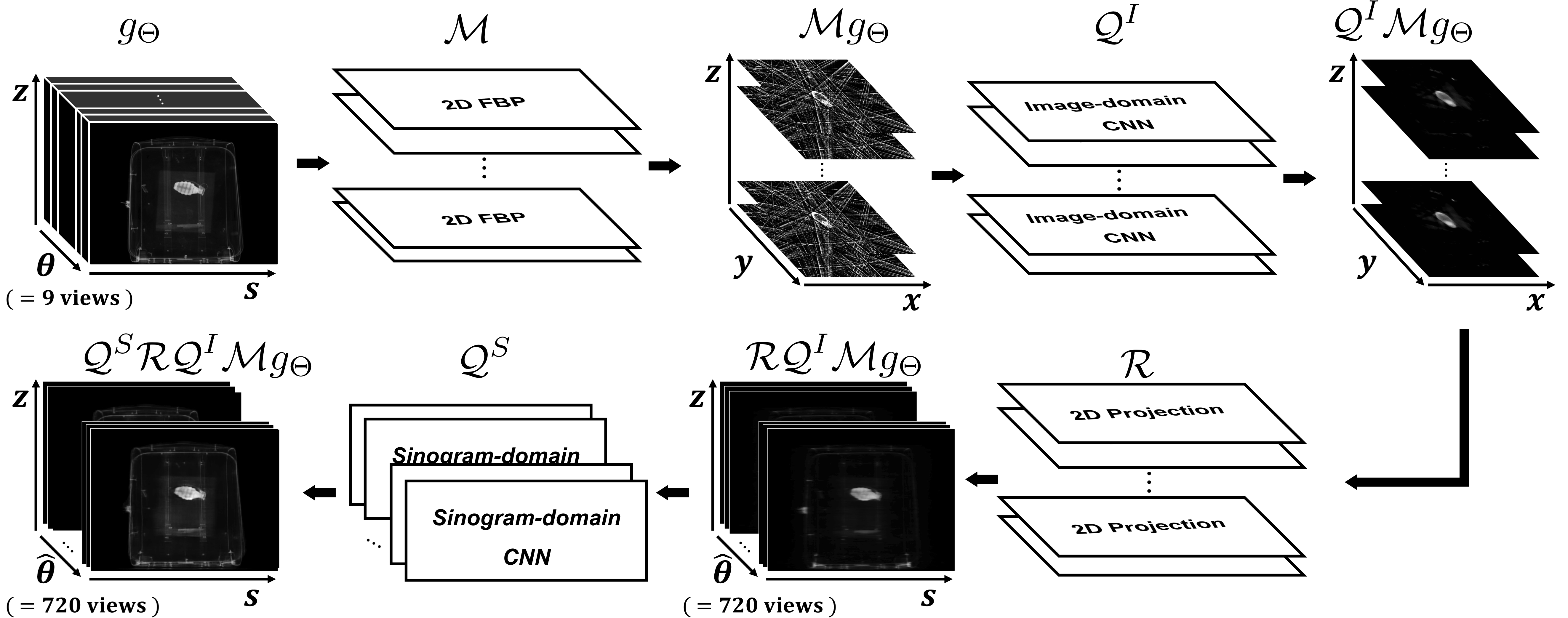}}
  \centerline{}\medskip
\vspace{-0.5cm}
\caption{Sinogram interpolation  flow for the proposed method. The final reconstruction is obtained by applying the FBP for the interpolated sinogram data.}\label{fig:flow} 
\end{figure*}

For sparse-view  CT EDS, model-based iterative reconstruction (MBIR) with the total variation (TV) penalty
have been extensively investigated  \cite{mandava2017image, kisner2012limited}.   
Inspired by the recent success of deep learning approach for sparse view and limited angle CT \cite{han2016deep,han2017framing,jin2017deep,gu2017multi} that outperform the classical MBIR approach, this paper aims at developing
a deep learning approach for real-world sparse view CT EDS. However, neural network training using the retrospective angular subsampling as  in the
existing works  \cite{han2016deep,han2017framing,jin2017deep,gu2017multi}  is not possible for our prototype system, since there are no ground-truth data for the real world sparse view CT EDS.
We therefore propose a novel deep learning approach composed of image domain and sinogram domain learning that compensate
for the imperfect label data.

%
%
%
%


\section{Theory}

\subsection{Problem Formulation}

Recall that the forward model for sparse view CT EDS system can be represented by
\begin{eqnarray}\label{eq:fwd}
g_\Theta  = \Pc_\Theta \Rc f
\end{eqnarray}
where $\Rc$ denotes the 3D projection operator from an $x-y-z$ volume image to a $s-\theta-z$ domain sinogram data with $s,\theta$ and $z$ denoting the detector,  projection angle,  
and the direction of the conveyor belt travel, respectively. See Fig.~\ref{fig:flow} for the coordinate systems.
In \eqref{eq:fwd}, $\Pc_\Theta$ denotes the view sampling operator for the measured angle set $\Theta$, and
$g_\Theta$ refers to the measured sinogram data.
For each projection view data, we use the notation $g_\theta$ and $\Pc_\theta$, where $\theta$ denotes the specific  view.

The main technical issue of the sparse view CT reconstruction is the non-uniqueness of the solution for \eqref{eq:fwd}.
More specifically, there exists a  null spacce  $\Nc_\Theta$ such that
$$\Pc_\Theta \Rc h=0,\quad \forall h \in \Nc_\Theta,$$
which leads to  infinite number of feasible solutions.
To avoid the non-uniqueness of the solution, constrained form of
the penalized MBIR can be formulated as : 
\begin{eqnarray}\label{eq:MBIR}
 \min_{f\in \Rd^3}  \|\mathrm{L} f\|_{1}, \quad \mbox{subject to} \quad g_\Theta = \Pc_\Theta\Rc  \  ,
\end{eqnarray}
where 
 $\mathrm{L}$ refers to a linear operator and  $\|\cdot\|_1$ denotes the $l_1$ norm.  For the case of the TV penalty,
$\mathrm{L}$ corresponds to the derivative. 
Then, the uniqueness of \eqref{eq:MBIR} is guaranteed that if the $\Nc_{\mathrm{L}} \cap \Nc_\Theta = \{0\}$,
where $\Nc_{\mathrm{L}}$ denotes the null space of the operator $\mathrm{L}$.

Instead of designing a linear operator ${\mathrm{L}}$ such that  the common null space of $\Nc_\Theta$ and $\Nc_{\mathrm{L}}$ to be zero,
we can design a frame $\Wc$,  its dual $\tilde \Wc$, and shrinkage operator $S_\lambda$ such that
$\tilde \Wc^\top \Wc=I$ and
$$\tilde \Wc^\top S_\lambda \Wc (f^*+g)  = f^*\quad \forall
g \in \Nc_\Theta$$
for the ground-truth image $f^*$.
This frame-based regularization is also an active field of research for image denoising,
inpainting, etc \cite{cai2008framelet}.
One of the most important contributions of the deep convolutional framelet theory \cite{ye2017deep}
is that $\Wc$ and $\tilde \Wc^\top$ correspond to the encoder and decoder structure of a convolutional neural network (CNN), respectively,
and the shrinkage operator $S_\lambda$ emerges by controlling the number of filter channels and nonlinearities.
More specifically,  a convolutional neural network  can be designed such that 
$\Qc =\tilde \Wc^\top S_\lambda \Wc $
and 
\begin{eqnarray}\label{eq:Qc1}
\Qc(f^*+h) = f^* ,\quad \forall h \in \Nc_\Theta \quad .
\end{eqnarray}
In other word, \eqref{eq:Qc1} directly removes the null space component.
Eq.~\eqref{eq:Qc1} is the constraint we use for training our neural network.


\subsection{Derivation of Image and Projection Domain CNNs}


More specifically, our sparse view  reconstruction algorithm  finds the unknown  $f\in \Rd^3$ that satisfy both data fidelity
and the so-called frame constraints \cite{ye2017deep}:
\begin{eqnarray}\label{eq:constraint}
g_\Theta  = \Pc_\Theta\Rc f ,\quad  \Qc^I (f) = f^*\quad , 
\end{eqnarray}
where $\Qc^I$ is the image domain CNN that satisfies \eqref{eq:Qc1}
and $f^*$ denotes the ground-truth images that are available for  training data.
Now, by defining $\Mc$ as a right-inverse of $\Pc_\Theta \Rc$, i.e. $(\Pc_\Theta\Rc)\Mc g_\Theta =g_\Theta, \forall g_\Theta$,
we have
$$ \Mc g_\Theta  = f^*+h $$
for some $h\in \Nc_\Theta$, since the right inverse is not unique due to the existence of the null space.
Thus,  we can show that $ \Mc g_\Theta$ is the feasible solution for 
\eqref{eq:constraint}, since we have 
\begin{eqnarray}\label{eq:Q2}
\Qc^I \Mc g_\Theta =  \Qc^I \left(f^*+h\right) =   f^*\quad ,
\end{eqnarray}
for the training data,
and
\begin{eqnarray}\label{eq:fidelity}
 \Pc_\Theta\Rc \Mc g_\Theta  =  \Pc_\Theta\Rc (f^*+h) = g_\Theta\quad.
 \end{eqnarray}
Therefore, the neural network training problem to satisfy \eqref{eq:constraint} can be equivalently
represented by
\begin{eqnarray}\label{eq:opt}
\min_{\Qc^I} \sum_{i=1}^N\|f^{*(i)} - \Qc^I \Mc g_\Theta^{(i)} \|^2
\end{eqnarray}
where $\{(f^{*(i)},g_\Theta^{(i)})\}_{i=1}^N$ denotes the training data set composed of ground-truth image an its sparse view  projection.
Since a representative right inverse for the sparse view projection is the inverse Radon transform after zero padding to the missing view,
 $\Mc g_\Theta^{(i)}$ in \eqref{eq:opt} can be implemented using the standard FBP algorithm.
 In fact, this is the main theoretical ground for the success of image domain CNN when the ground-truth data is available
 \cite{han2016deep,han2017framing,jin2017deep,gu2017multi}.
 Moreover, the fan-beam rebinning makes the problem separable for each $z$ slices, so we use the 2D FBP for each slice  as shown in Fig.~\ref{fig:flow}.

However, the main technical difficulties in our 9-view CT EDS system is that we do not have 
ground-truth image $\{f^{*(i)}\}_{i=1}^N$.  One could use physical phantoms and atomic number to form a set of ground-truth
images, but those data set may be different from the real carry-on bags, so we need a new method to account for
the lack of ground-truth for neural network training.
Thus,  to overcome the lack of the ground-truth data,
the approximate label images are generated using an MBIR with TV penalty.
Then, using MBIR reconstruction as label data $\{f^{*(i)}\}_{i=1}^N$,
an 2D image domain network $\Qc^I$ is trained to learn the mapping between the  artifact-corrupted 2D image and MBIR reconstruction  in $x-y$ domain.

One downside of this approach is that the network training by \eqref{eq:opt} is no more optimal, since the label data is
not the ground-truth image. Thus, the generated sinogram data from the denoised 3D volume may be biased.
Thus,  
we impose  additional frame constraint to the sinogram data in addition to \eqref{eq:constraint}: 
\begin{eqnarray}\label{eq:constraint2}
 g_\theta^* = \Qc^S\left( g_\theta\right) , 
\end{eqnarray}
for the measured angle $\theta$, where $\Qc^S$ is the $s-z$ sinogram domain CNN and $g_\theta^*$ denotes the ground-truth sinogram
data measured at $\theta$.
Then, Eq.~\eqref{eq:constraint2} leads to the following network training:
\begin{eqnarray}\label{eq:opt2}
\min_{\Qc^S} \sum_{\theta\in \Theta}\sum_{i=1}^N \|g_\theta^{*(i)} - \Qc^S \left(\Pc_\theta \Rc \Qc^I\Mc g_\Theta^{(i)}\right) \|^2
\end{eqnarray}
More specifically, as shown in Fig.~\ref{fig:flow},
3D sinogram data   is generated in the $s-\theta-z$ domain by applying the forward projection operator along 720-projection views
after  stacking the image domain network output over multiple slices to form 3D reconstruction volume in the $x-y-z$ domain. 
Next, a 2D sinogram domain network $\Qc^S$ is trained so that it can learn the mapping between the synthetic $s-z$ sinogram data 
and the real projection data in the $s-z$ domain.  Since the real projection data is available only  in 9 views, this sinogram network
training is performed using synthetic and real projection data in the measured projection views.
The optimization problems \eqref{eq:opt} and \eqref{eq:opt2} can be solved sequentially or simultaneously,
and in this paper we adopt the sequential optimization approach for simplicity.

After the neural networks $\Qc^I$ and $\Qc^S$ are trained, the inference can be done simply by obtaining
$x-y-z$ volume images from the 9 view projection data by slice-by-slice FBP algorithm, which are then
fed into $\Qc^I$ to obtain the denoised 3D volume data.  Then,  by applying projection operator, we generate
720 projection view data in $s-\theta-z$ domain,  which are fed into the $\Qc^S$ to obtain denoised sinogram data
for each $\theta$ angle. Then, the final reconstruction is obtained by applying FBP algorithms.
One could use post-processing using additional TV-based denosing.
This algorithmic flow is illustrated in Fig.~\ref{fig:flow}.

\section{Methods}

\subsection{Real CT EDS data Acquisition}

We collected CT EDS data using the prototype stationary 9 view dual energy CT-EDS system developed by GEMSS Medical Systems, Korea
as shown in 
 Fig. \ref{fig:ct_system}. 
the distance from source to detector (DSD) and the distance from source to object (DSO) are $1202.6 \rm{mm}$ and $648.2 \rm{mm}$, respectively. The number of detector is $384$ with a pitch of $1.5 \rm{mm}$. The region of interest (ROI) is $256\times256$ and the pixel size is $2 \rm{mm}^2$. 
The detectors collect low and high energy X-ray at 80KVp and 120KVp, respectively.

We collect 47 sets of projection data from the  prototype CT EDS baggage scanner. Among  the 47 sets, 32 dataset are simple-objects and the other set are realistic carry-on bags. The 47 set of 28 simple- and 13 baggage-objects was used during the training phase,  and the  validation was performed by two simple- and one baggage-object. The other set was used for test.

\subsection{Network Architecture and Training }

\begin{figure}[t]
  \centering
  \centerline{\includegraphics[width = 0.5\textwidth]{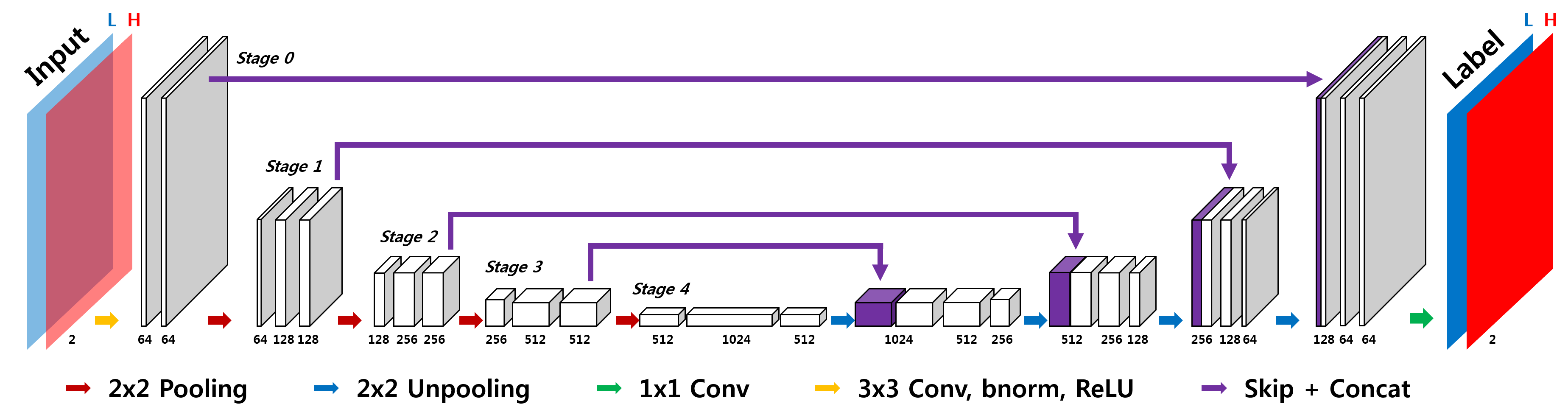}}
  \centerline{}\medskip
\vspace{-1cm}
\caption{CNN architecture for our image and singoram domain networks. }\label{fig:network} 
\end{figure}

Fig. \ref{fig:network} illustrates modified the U-Net  structure \cite{ronneberger2015u} for the image domain and the sinogram domain networks.
To account for the multi-energy image and sinogram data, the input for the network is two channel multi-energy image and sinogram data. 
The proposed network consists of convolution layer, batch normalization, rectified linear unit (ReLU) \cite{krizhevsky2012imagenet}, and contracting path connection with concatenation \cite{ronneberger2015u}. A detail parameters are illustrated as shown in Fig. \ref{fig:network}.

The proposed networks were trained by stochastic gradient descent (SGD). The regularization parameter was $\lambda = 10^{-4}$. The learning rate has been set from $10^{-3}$ to $10^{-5}$, which has been reduced step by step in each epoch. The number of epoch was 200. The batch size was 12 and the patch size for image and projection data are $256\times256\times2$ and $768\times384\times2$, respectively. 
The network was implemented using MatConvNet toolbox (ver.24) \cite{vedaldi2015matconvnet} in the MATLAB 2015a environment (MathWorks, Natick). Central processing unit (CPU) and graphic processing unit (GPU) specification are i7-7700 CPU (3.60 GHz) and GTX 1080 Ti GPU, respectively.

\section{Experimental Results}

\begin{figure}[t]
  \centering
  \centerline{\includegraphics[width =8cm,height=9.5cm]{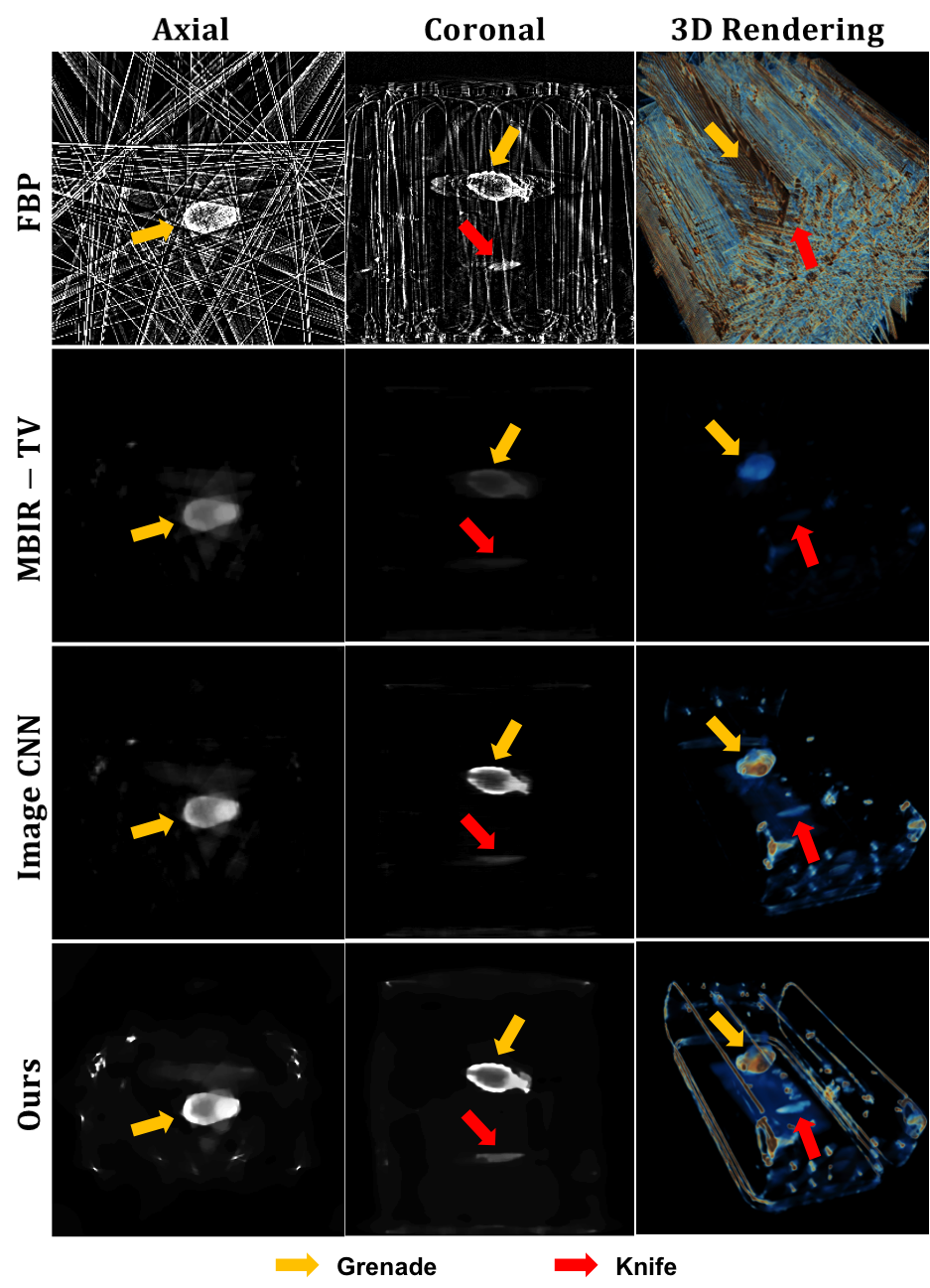}}
  \centerline{}\medskip
\vspace{-1cm}
\caption{Reconstruction results by various methods from  9-views  CT-EDS..  }\label{fig:result_pht} 
\end{figure}

%


To evaluate the performance of the proposed method, we  perform image reconstruction from real 9-view CT EDS prototype system.
Fig. \ref{fig:result_pht} illustrates image reconstruction results of bag using various methods such as FBP,  MBIR with TV penalty,
image domain CNN  \cite{han2016deep,jin2017deep}, and the proposed method.  The FBP reconstruction results suffered from severe streaking
artifacts, so it was difficult to see the threats in the tomographic reconstruction and 3D rendering.
The MBIR and image domain CNN were slight better in their reconstruction quality, but the detailed 3D structures were not fully
recovered and several objects were not detected as indicated by the red arrow  in Fig. \ref{fig:result_pht}.
Moreover, the 3D rendering results in  Fig. \ref{fig:result_pht} correctly identify the shape of  grenade and knife as well as the frame of the bag,
which was not possible using other methods.

\begin{figure}[!hbt]
  \centering
  \centerline{\includegraphics[width = 0.4\textwidth]{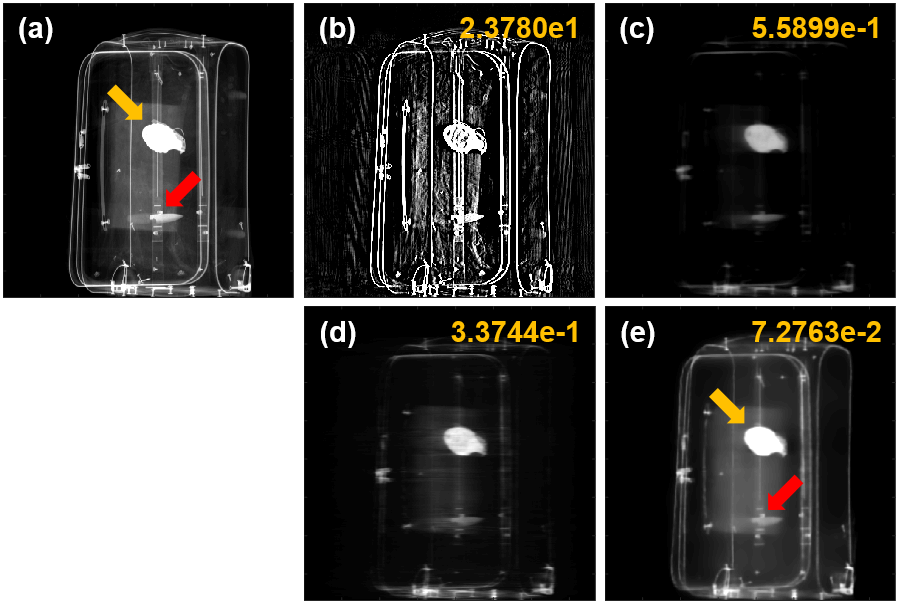}}
  \centerline{}\medskip
\vspace{-0.8cm}
\caption{A $s-z$ domain sinogram data from (a) measurement,  (b) 9-views FBP (c) MBIR, (d) image CNN, and (e) the proposed method. The number written in the images is the NMSE value. Yellow and red arrows indicate grenade and knife, respectively.}\label{fig:result_r_pht} 
\end{figure}

Because we do not have the ground-truth in the image domain,
we perform quantitative evaluation using normalized mean squares error (NMSE) in the sinogram domain. More specifically,
after obtaining the final reconstruction, we perform the forward projection to generate the sinogram data
in the measured projection view and calculated the normalized mean square errors.
Table~\ref{tab_quality} showed that the proposed method provides the most accurate sinogram data compared to the
other methods. Moreover, the $s-z$ projection data in Fig.~\ref{fig:result_r_pht} showed that
the projection data from the proposed method is much closer to the ground-truth measurement data.

\begin{table}[h!] 
\caption{NMSE value comparison of various  methods.}
\vspace*{-0.5cm}
\label{tab_quality}
\begin{center}
\begin{tabular}{c|cccc}
\hline
Energy level &  FBP &  MBIR-TV  & Image CNN &  Ours \\
\hline\hline
80 kvP &
1.6647e+1 &
5.8247e-1 &
3.3207e-1&
0.6845e-1\\
120 kvP&
1.0536e+1 &
6.0440e-1 &
3.2249e-1 &
0.5450e-1 \\
\hline
\end{tabular}
\end{center}
\end{table}

%

\section{Conclusion}

In this paper, we proposed a novel deep learning reconstruction algorithm for a prototype 9-view dual energy  CT EDS for carry-on baggage scanner. 
Even though the number of projection view was not sufficient for high equality 3D reconstruction,
our method learns the relationships between the 2D tomographic slices in $x-y$ domain  as well as the 2D projections in $s-z$ domain such that
the artifact-corrupted image and sinogram data can be successively refined to obtain high quality images. 
Using real data from our prototype 9-view CT EDS system, we demonstrated
that  the proposed method outperforms the existing algorithms, delivering high quality three reconstruction for threat detection.


\section*{Acknowledgment}

This work is supported by Korea Agency for Infrastructure Technology Advancement, Grant number
17ATRP-C071164-05-000000.


\end{document}